\documentclass[10pt,twocolumn,letterpaper]{article}

\usepackage{times}
\usepackage{epsfig}
\usepackage{amsmath}
\usepackage{amssymb}

\usepackage{float}

\usepackage[pagenumbers]{cvpr} 

\usepackage{graphicx}
\usepackage{amsmath}
\usepackage{amssymb}
\usepackage{booktabs}

\usepackage{times}
\usepackage{epsfig}
\usepackage{graphicx}
\usepackage{amsmath}
\usepackage{amssymb}
\usepackage{bm}
\usepackage{multirow}
\usepackage{dsfont}
\usepackage{xcolor}
\usepackage{wrapfig}
\usepackage{booktabs}
\usepackage{color}
\usepackage{verbatim}
\usepackage{caption}
\usepackage{ulem}
\usepackage{enumitem}
\usepackage{placeins}
\usepackage{ragged2e}
\usepackage{diagbox}
\usepackage{enumitem}

\usepackage{epsfig}
\usepackage{textcomp}
\usepackage{bm}
\usepackage{soul}
\usepackage{makecell}

\usepackage{makecell}
\usepackage{arydshln}
\usepackage{diagbox}
\usepackage{ulem}

\usepackage{booktabs}

\useunder{\uline}{\ul}{}

\newcommand{\ieours}{\textit{i}.\textit{e}., }
\newcommand{\egours}{\textit{e}.\textit{g}., }

\newcommand{\nickname}{method}

\newcommand{\TBD}[1]{\textcolor{black}{#1}}
\newcommand{\qy}[1]{\textcolor{black}{#1}}

\usepackage[pagebackref,breaklinks,colorlinks]{hyperref}

\usepackage[capitalize]{cleveref}
\crefname{section}{Sec.}{Secs.}
\Crefname{section}{Section}{Sections}
\Crefname{table}{Table}{Tables}
\crefname{table}{Tab.}{Tabs.}

\def\cvprPaperID{4440} 
\def\confName{CVPR}
\def\confYear{2022}

\begin{document}




\title{\TBD{3DAC: Learning Attribute Compression for Point Clouds}}




\author{Guangchi Fang\textsuperscript{1,2}, Qingyong Hu\textsuperscript{3}, Hanyun Wang\textsuperscript{4}, Yiling Xu\textsuperscript{5}, 
Yulan Guo\textsuperscript{1,2,6}\thanks{Corresponding author: Yulan Guo (guoyulan@sysu.edu.cn).} \\
\textsuperscript{1}Sun Yat-sen University, \textsuperscript{2}The Shenzhen Campus of Sun Yat-sen University, \textsuperscript{3}University of Oxford\\
\textsuperscript{4}Information Engineering University, \textsuperscript{5}Shanghai Jiaotong University\\ \textsuperscript{6}National University of Defense Technology}

\maketitle

\begin{abstract}
We study the problem of attribute compression for large-scale unstructured 3D point clouds. Through an in-depth exploration of the relationships between different encoding steps and different attribute channels, we introduce a deep compression network, termed 3DAC, to explicitly compress the attributes of 3D point clouds and reduce storage usage in this paper. Specifically, the point cloud attributes such as color and reflectance are firstly converted to transform coefficients. We then propose a deep entropy model to model the probabilities of these coefficients by considering information hidden in attribute transforms and previous encoded attributes. Finally, the estimated probabilities are used to further compress these transform coefficients to a final attributes bitstream. Extensive experiments conducted on both indoor and outdoor large-scale open point cloud datasets, including ScanNet and SemanticKITTI, demonstrated the superior compression rates and reconstruction quality of the proposed \nickname{}.


\end{abstract}


\section{Introduction}
\label{sec:intro}

\begin{figure}[t]
	\centering
	\includegraphics[width=1.0\linewidth]{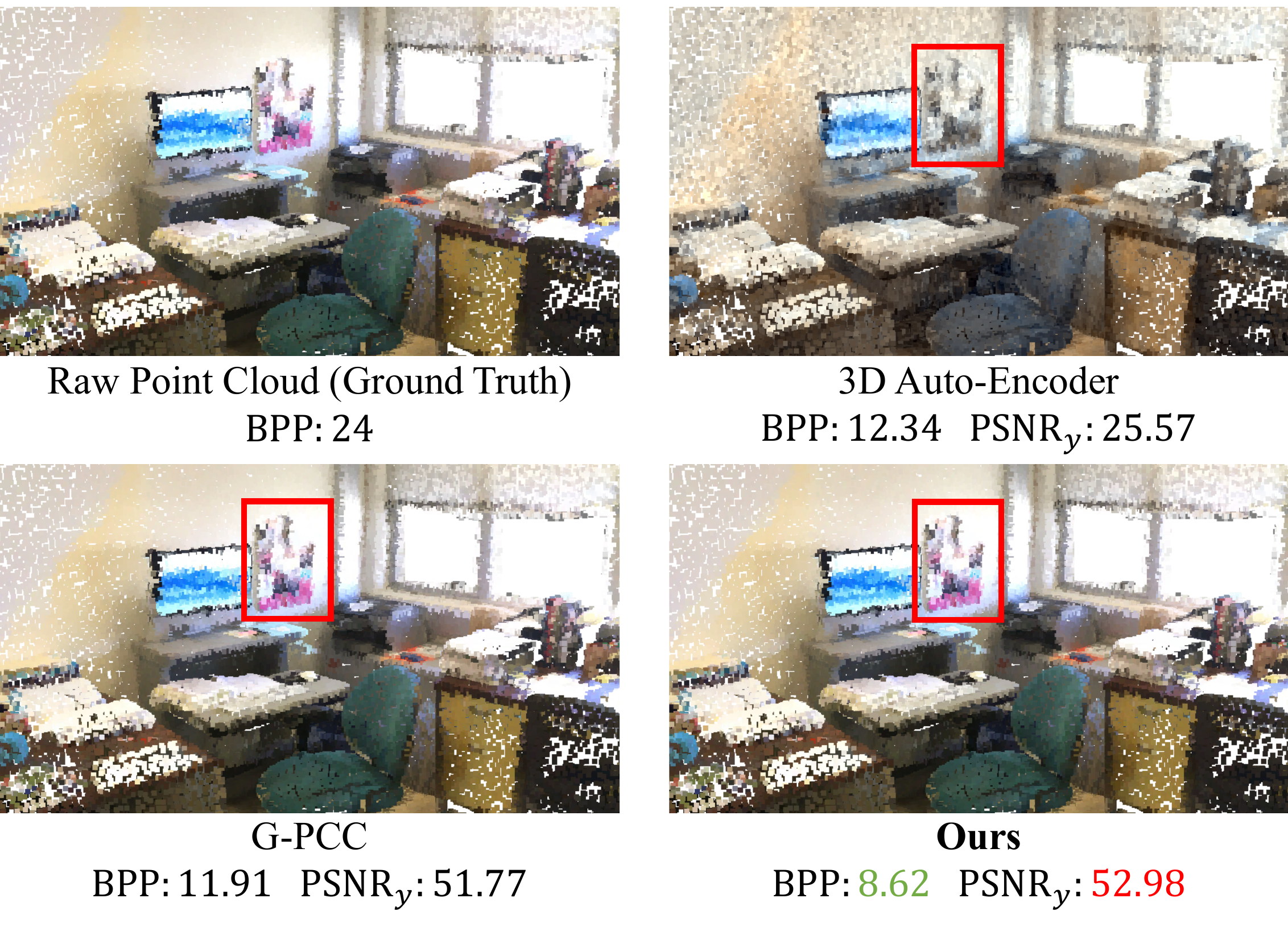}
	
    \caption{Qualitative point cloud attribute compression results of 3D auto-encoder, G-PCC, and our \nickname{} on the ScanNet \cite{scannet}. Bits Per Point (BPP) and Peak Signal-to-Noise Ratio (PSNR) of the luminance component are reported. Note that, the raw point clouds usually use uint8 RGB values (\textit{i.e.,}  $8\times3=24$ BPP).}
	\label{fig:teaser}
\end{figure}

	


As a common 3D data representation, point clouds have been widely used in a variety of real applications such as mixed reality \cite{mix_reality}, self-driving vehicles \cite{geiger2012we, hu2021learning}, and high-resolution mapping \cite{mapping, hu2022sensaturban}. Thanks to the remarkable progress achieved in 3D acquisition, point clouds become increasingly accessible. However, the storage and transmission of massive irregularly sampled points pose a new challenge to existing compression techniques. In particular, along with the raw 3D coordinates of points, the compression of their attributes (\textit{e.g.}, color, reflectance) is also non-trivial\footnote{For example, with the standard point cloud compression algorithm of MPEG (\textit{i.e.}, G-PCC \cite{schwarz2018emerging}), attribute compression takes up to 60\% and 90\% of the overall bitstream for lossy and lossless compression, respectively.}. In this regard, we will study effective attribute compression for unstructured point clouds in this paper.

To achieve point cloud attribute compression, early works \cite{zhang2014point, thanou2015graph, shao2017attribute, shao2018hybrid, de2016compression, souto2020predictive, pavez2021multi, pavez2020region} usually apply image processing techniques to 3D point clouds. In particular, early methods usually follow a two-step framework, \textit{i.e.}, initial coding of attributes and entropy coding of transform coefficients. A number of approaches focus on developing sophisticated initial coding algorithms, which convert point cloud attributes to coefficients in a specific domain. However, entropy coding, which further losslessly encodes coefficients to a final bitstream, has been largely overlooked. There are only a few entropy coders \cite{malvar2006adaptive, gu20193d, schwarz2018emerging} have been proposed for attribute compression in recent works. In general, entropy coding is independently included in the aforementioned attribute compression framework with initial coding. More specifically, these traditional hand-crafted entropy coders take coefficients as a sequence of symbols and estimate the probability distribution\footnote{According to the information theory \cite{shannon1948mathematical}, a lower bound of bitrate can be achieved by an entropy coder given the actual distribution of the transmitted coefficients. } only considering the previous input symbols as context information. Most of prior entropy coders do not incorporate geometry information of point clouds or context information of initial coding. Moreover, these approaches separately encode attributes for each channel and do not make full use of the inter-channel correlations between different attributes.


In this paper, we propose a learning-based compression framework, termed 3DAC, for point cloud attributes. Specifically, the proposed 3DAC adopts Region Adaptive Hierarchical Transform (RAHT) \cite{de2016compression} for initial coding. Then, we propose an attribute-oriented deep entropy model to estimate the probability distribution of transform coefficients. In particular, we model the probabilities of these coefficients by exploring context information from the initial coding stage and inter-channel correlations between different attributes. As shown in Fig. \ref{fig:teaser}, our method achieves higher reconstruction quality with a lower bitrate, indicating excellent attribute compression performance. The main contributions of this paper are as follows:



\vspace{-0.2cm}
\begin{itemize}
\setlength{\itemsep}{0pt}
\setlength{\parsep}{0pt}
\setlength{\parskip}{0pt}
    \item \TBD{We introduce a learning-based, effective framework for attribute compression of 3D point clouds, with competitive compression performance.}
    \item We propose an attribute-oriented deep entropy model to connect the initial coding and entropy coding steps in attribute compression, and explore the inter-channel correlations between different attributes.
    \item We demonstrate the state-of-the-art compression performance of the proposed \nickname{} on both indoor and outdoor point cloud datasets.
\end{itemize}

\section{Related Work}
\label{sec:related_work}

\begin{figure*}[thb]
	\centering
	\includegraphics[width=1.0\linewidth]{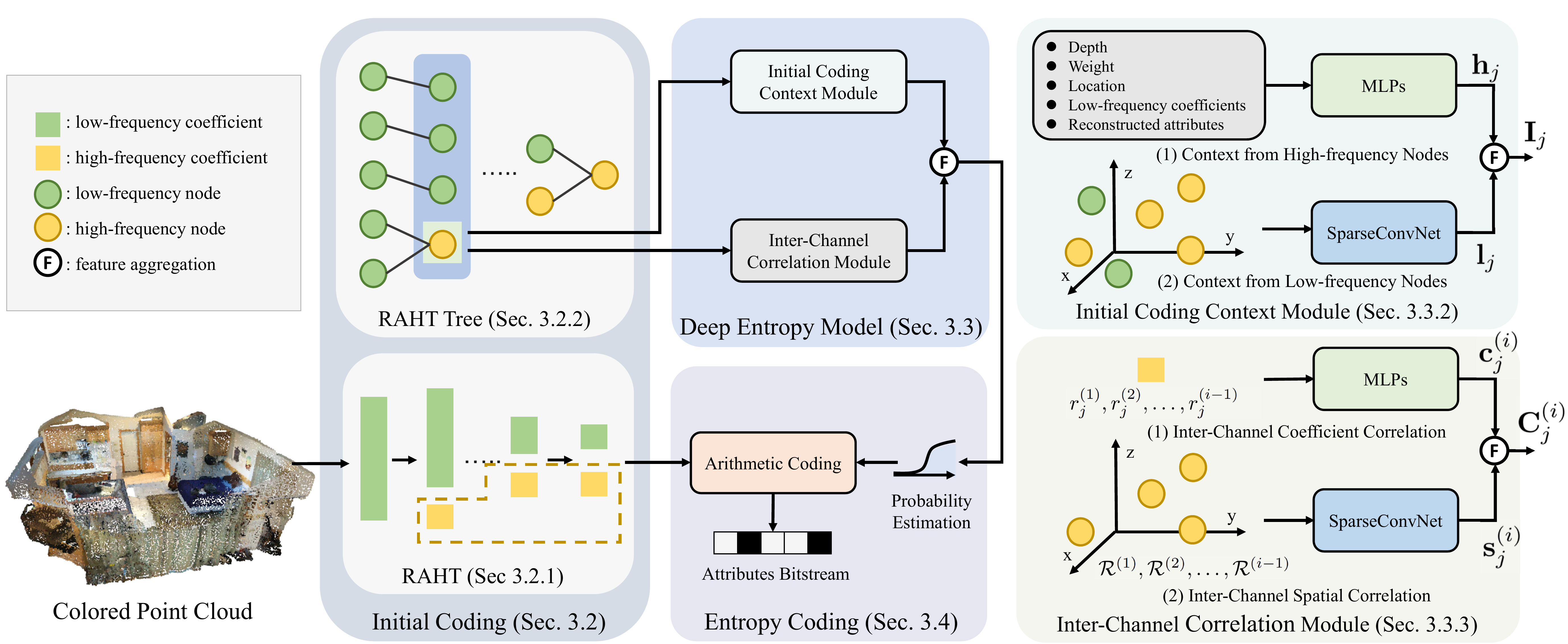}
	
    \caption{The network architecture of our point cloud attribute compression method. }
	\label{fig:method_overview}
\end{figure*}

\subsection{Point Cloud Geometry Compression}


Point cloud geometry compression aims at compressing the 3D coordinates of points. In light of the unstructured nature of point clouds, existing methods  \cite{meagher1982geometric, schwarz2018emerging} usually leverage an advanced data structure such as octree to organize the raw point clouds. For example, G-PCC \cite{schwarz2018emerging} includes an octree-based method for geometry compression. Later, in several learning-based approaches \cite{huang2020octsqueeze, biswas2020muscle, que2021voxelcontext}, the octree structure was also used for initial coding and octree-structured entropy models were proposed for probability estimation. These octree-based approaches are mainly tailored for geometry compression. In addition, deep image compression \cite{balle2018variational} are   extended  to the 3D domain \cite{huang20193d, yan2019deep, quach2019learning,  quach2020improved, gao2021point, wang2021multiscale} with 3D auto-encoders \cite{qi2017pointnet, brock2016generative, choy20194d}. However, due to the high complexity of point cloud attributes, neural networks tend to ignore high-frequency components (\textit{i.e.}, attribute details). Therefore, it is infeasible to extend this framework to attribute compression.

\subsection{Point Cloud Attribute Compression}

The objective of attribute compression is to compress the attributes (\textit{e.g.}, color, reflectance) of points. The general idea is to apply traditional image compression techniques to 3D point clouds, which usually includes initial coding of attributes and entropy coding of transform coefficients.

Initial coding aims at capturing the signal redundancy in the transform domain. For example, Zhang et al. \cite{zhang2014point} first split a point cloud into several blocks through octree and used graph transform to convert attributes to eigenvectors. Shao et al. \cite{shao2017attribute, shao2018hybrid} organized a point cloud with an KD-Tree and adopted graph transform for attribute compression. Due to the repeated use of eigendecomposition, graph transform-based methods are usually time-consuming. To handle this problem, Queiroz et al. \cite{de2016compression} proposed a variation of the Haar wavelet transform, namely, region adaptive hierarchical transform (RAHT), for real-time point cloud transmission. In several follow up works, RAHT is extended with intra-frame and inter-frame prediction \cite{souto2020predictive, pavez2021multi}, graph transform \cite{pavez2020region}, and fixed-point operations \cite{sandri2019integer}.


Entropy coding tends to encode symbols into a bitstream with the estimated distribution through an entropy coder, such as Huffman \cite{huffman1952method}, arithmetic \cite{witten1987arithmetic}, and Golomb-Rice coders \cite{weinberger2000loco, richardson2004h}. Zhang et al. \cite{zhang2014point} and Queiroz et al. \cite{de2017transform} assumed that transform coefficients follow a certain Laplacian distribution and then encoded them with arithmetic coding. However, the estimated distribution is usually difficult to approximate the real distribution due to the naive assumption. In \cite{de2016compression, souto2020predictive, pavez2021multi}, an adaptive run-length Golomb-Rice (RLGR) coder \cite{malvar2006adaptive} was adopted for entropy coding. Two variations of run-length coding were also adopted by \cite{schwarz2018emerging, gu20193d}. All these entropy coders take input symbols as sequential data and only adjust their encoding parameters depending on the previous input symbols.

A handful of recent works \cite{quach2020folding, biswas2020muscle, sheng2021deep} started to explore deep learning techniques for attribute compression. Maurice et al. \cite{quach2020folding} used FoldingNet \cite{yang2018foldingnet} to reorganize a point cloud as an image and then compressed the image with a conventional 2D image codec. Muscle \cite{biswas2020muscle} is proposed for dynamic LiDAR intensity compression, which exploited the LiDAR geometry and intensity of the previous frame for its deep entropy model. The compression performance can further be improved by exploiting context information from other initial coding methods and point cloud data structures.

\subsection{Deep Image Compression}

Traditional image compression methods \cite{wallace1992jpeg, goyal2001theoretical, skodras2001jpeg, sullivan2012overview, bpg} typically contain three steps: transformation, quantization, and entropy coding. In particular, the source image is first transformed from signal domain to frequency domain to capture spatial redundancy with a few transform coefficients. These coefficients are then quantized with a hand-crafted quantization table and entropy coded for further compression. 



In contrast to traditional image compression approaches, deep learning based methods jointly optimize the three steps. Most recent methods \cite{balle2017end, balle2018variational, yang2021slimmable, he2021checkerboard, deng2021deep, gao2021neural, bai2021learning} follow an auto-encoder pipeline. These methods formulate a nonlinear function with an encoder to map the source image to a more compressible latent space, and recover the image from the latent codes with a decoder. Then, they also model the entropy of the latent codes with neural networks. For example, Ball{\'e} et al. \cite{balle2017end} used 2D convolution neural networks to model both the non-linear transform and the entropy model for image compression.



\section{Methodology}
\label{sec:method}




\subsection{Overview}

Given a 3D point cloud, its geometry is assumed to have been transmitted separately and this paper mainly focuses on the task of point cloud attribute compression. We propose a learning based attribute compression framework, termed 3DAC, to reduce the storage while ensuring reconstruction quality. Without loss of generality, our framework takes a colored point cloud as its input, as shown in Fig. \ref{fig:method_overview}. Note that, other attributes, such as opacity and reflectance, can also be compressed with our framework.


As depicted in Fig. \ref{fig:method_overview}, we first adopt an efficient initial coding method (\ieours RAHT) to decompose point cloud attributes into low- and high-frequency coefficients. Then, we model this process with a tree structure (\ieours RAHT tree). The resulting high-frequency coefficients are quantized and formulated as a sequence of symbols via the RAHT tree. Next, we propose an attribute-oriented entropy model to exploit the initial coding context and the inter-channel correlation. Consequently, the probability distribution of each symbol can be well modelled. In the entropy coding stage, the symbol stream and the predicted probabilities are further passed into an arithmetic coder to produce the final attributes bitstream.



\subsection{Initial Coding}\label{sec:Initial Coding}

The initial coding methods are capable of capturing spatial redundancy of point cloud attributes by converting attributes to transform coefficients. Here, we adopt Region Adaptive Hierarchical Transform (RAHT) \cite{de2016compression} for initial coding due to its simplicity and flexibility. Note that, it is possible to integrate an improved version of RAHT into our framework for better performance.



\subsubsection{Region Adaptive Hierarchical Transform}\label{raht}

RAHT is a variation of Haar wavelet transform, which converts point cloud attributes to frequency-domain coefficients. Specifically, the raw point cloud is first voxelized, and the attributes are then decomposed to low- and high-frequency coefficients by combining points from individual voxels to the entire 3D space. Here, we briefly revisit RAHT \cite{de2016compression} through a 2D toy example.

\begin{figure}[t]
	\centering
	\includegraphics[width=1.0\linewidth]{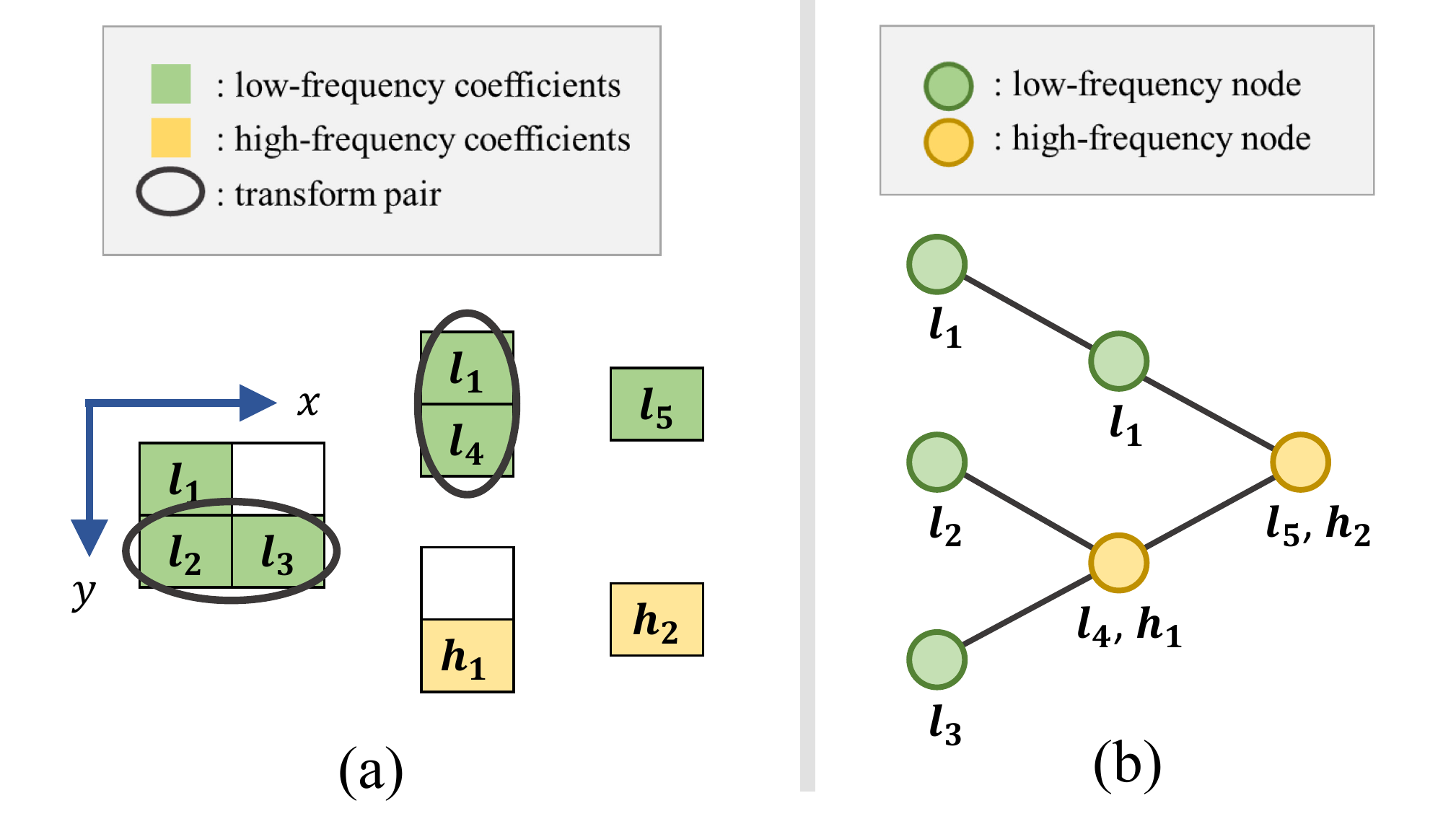}
	
    \caption{An illustration of RAHT and RAHT tree. (a) A 2D example of RAHT, (b) the corresponding RAHT tree.}
	\label{fig:method_raht}
\end{figure}

Figure \ref{fig:method_raht}(a) shows an 2D example of RAHT. Only two dimensions (\textit{i.e.}, $x$ and $y$) are considered in this case. In Fig. \ref{fig:method_raht}(a), $l_{i}$ and $h_{i}$ denote low- and high-frequency coefficients, respectively. In the first block of Fig. \ref{fig:method_raht}(a), points are represented as occupied voxels, and the attributes of corresponding points are denoted as low-frequency coefficients $l_{1}$, $l_{2}$ and $l_{3}$. In the encoding stage, we apply transform along the $x$ axis first and then the $y$ axis by turn until all voxels are merged to the root space. At the first depth level, $l_{2}$ and $l_{3}$ are transformed to $l_{4}$ and $h_{1}$, while $l_{1}$ is transmitted to the next depth level directly due to lack of neighbor along the $x$ axis. In the second depth level, $l_{1}$ and $l_{4}$ are transformed to $l_{5}$ and $h_{2}$ along the $y$ axis. During decoding, the DC coefficient $l_{5}$ and high-frequency coefficient $h_{2}$ are used to restore low-frequency coefficients $l_{1}$ and $l_{4}$. Similarly, $h_{1}$ is used for $l_{2}$ and $l_{3}$ with the reconstructed $l_{4}$. Thus, only the DC coefficient $l_{5}$ and all high-frequency coefficients $h_{1}$ and $h_{2}$ are required to be transmitted as symbols. 
That is, these coefficients have to be encoded for attribute compression.



\textcolor{black}{For 3D point clouds, RAHT transforms attributes to coefficients along three dimensions repeatedly (\egours along the $x$ axis first, then the $y$ axis and the $z$ axis) until all subspaces are merged to the entire 3D space. }
Two neighboring points are merged with the following transform:
\begin{equation}
\left[\begin{array}{l}
l_{d+1, x, y, z} \\
h_{d+1, x, y, z}
\end{array}\right]=\mathbf{T}_{w_{1}, w_{2}}\left[\begin{array}{c}
l_{d, 2 x, y, z} \\
l_{d, 2 x+1, y, z}
\end{array}\right]\label{con:raht},
\end{equation}
where $l_{d, 2 x, y, z}$ and $l_{d, 2 x+1, y, z}$ are low-frequency coefficients of two neighboring points along the $x$ dimension, and $l_{d+1, x, y, z}$ and $h_{d+1, x, y, z}$ are the decomposed low-frequency and  high-frequency coefficients. Here, $\mathbf{T}_{w_{1}}$ is defined as
\begin{equation}
\mathbf{T}_{w_{1}, w_{2}}=\frac{1}{\sqrt{w_{1}+w_{2}}}\left[\begin{array}{cc}
\sqrt{w_{1}} & \sqrt{w_{2}} \\
-\sqrt{w_{2}} & \sqrt{w_{1}}
\end{array}\right],
\end{equation}
where $w_{1}$ and $w_{2}$ are the weights (i.e., the number of leaf nodes) of $l_{d, 2x, y, z}$ and $l_{d, 2x+1, y, z}$, respectively. Low-frequency coefficients are directly passed to the next level if the point does not have a neighbor.

\subsubsection{RAHT Tree}\label{sec:raht_tree}

We model RAHT with a tree structure for context feature extraction in Sec. \ref{Our Deep Entropy Model}. In general, points are organized as a RAHT tree according to the hierarchical transform steps. Here, we show a constructed RAHT tree of the aforementioned 2D toy example in Fig. \ref{fig:method_raht}(b). Consistent to Sec. \ref{raht}, tree nodes of $l_{2}$ and $l_{3}$ are merged to that of $l_{4}$ and $h_{1}$, while $l_{1}$ and $l_{4}$ are merged to $l_{5}$ and $h_{2}$. In this RAHT tree, leaf nodes represent original voxelized points and internal nodes represent the corresponding subspace. In particular, the low-frequency coefficient of a RAHT tree node is passed to its parent node if this node has no neighbor along the transform direction. Otherwise, two nodes are merged to their parent. Meanwhile, low- and high-frequency coefficients are generated. 

For a better illustration of probability and context modeling in Sec. \ref{Our Deep Entropy Model}, we denote nodes only containing low-frequency coefficients as low-frequency nodes (\ieours green nodes in Fig. \ref{fig:method_raht}(b)), and others containing both low- and high-frequency coefficients as high-frequency nodes (\ieours yellow nodes in Fig. \ref{fig:method_raht}(b)).


\subsubsection{Serialization}\label{Serialization}

We quantize all high-frequency coefficients and serialize these quantized coefficients into a symbol stream with a breadth-first traversal of the RAHT tree. \TBD{Take the aforementioned 2D toy case as an example, we serialize coefficients of the root level as $\{h_{2}\}$ and the second level as $\{h_{1}\}$.} The serialization of coefficient is lossless and the attribute distortion only comes from the quantization step. Note that, all context information adopted in Secs. \ref{Initial Coding based Context Module} and  \ref{Inter Channel based Context Module} is accessible to our deep entropy model during entropy decoding given the point cloud geometry and the breadth-first traversal format. For the DC coefficient, we transmit it directly.


\subsection{Our Deep Entropy Model}\label{Our Deep Entropy Model}


According to the information theory \cite{shannon1948mathematical}, given the actual distribution of the transmitted high-frequency coefficients $\mathcal{R}$, a lower bound of bitrate can be achieved by entropy coding. However, the actual distribution $p(\mathcal{R})$ is usually unavailable in practice. To deal with this problem, we propose a deep entropy model to approximate the unknown distribution $p(\mathcal{R})$ with the estimated distribution $q(\mathcal{R})$. \TBD{In general, we first formulate $q(\mathcal{R})$ with initial coding context and inter-channel correlation, and then utilize a deep neural network (which consists of two modules) to approximate $q(\mathcal{R})$ with $p(\mathcal{R})$ using the cross-entropy loss.}


\subsubsection{Formulation}


Given a 3D point cloud and its associated RAHT tree with $m$ high-frequency nodes and $n$ attribute channels, the transmitted high-frequency coefficients are denoted as $\mathcal{R}=\left\{\mathcal{R}^{(1)}, \ldots, \mathcal{R}^{(n)}\right\}$ and $\mathcal{R}^{(i)}=\left\{r_{1}^{(i)}, \ldots, r_{m}^{(i)}\right\}$. Considering the inter-channel correlation, we first factorize $q(\mathcal{R})$ into a product of conditional probabilities:

\begin{equation}
q(\mathcal{R})=\prod_{i} q\left(\mathcal{R}^{(i)} \mid \mathcal{R}^{(i-1)}, \ldots, \mathcal{R}^{(1)}\right).
\end{equation}
\qy{Here, the probability distribution of $\mathcal{R}^{(i)}$ is assumed to depend on previous encoded sequences $\left\{\mathcal{R}^{(1)}, \ldots, \mathcal{R}^{(i-1)}\right\}$. It is noticed that the probability distribution $q(\mathcal{R})$ also depends on the context information from the initial coding stage, hence we further factorize $q(\mathcal{R}^{(i)})$ with initial coding context $\mathbf{I}_j$ of the coefficient $r_{j}^{(i)}$:
}
\begin{equation}
q(\mathcal{R}^{(i)})=\prod_{j} q\left(r_{j}^{(i)} \mid \mathbf{I}_j, \mathcal{R}^{(i-1)}, \ldots, \mathcal{R}^{(1)}\right).
\end{equation}
\TBD{Then, we model the estimated probability distribution $q\left(\cdot \mid \mathbf{I}_j, \mathcal{R}^{(i-1)}, \ldots, \mathcal{R}^{(1)}\right)$ through a probability density model \cite{balle2018variational} with two proposed context modules, including the initial coding context module (which encodes the information from attribute transform) and the inter-channel correlation module (which explores the dependence on previously encoded attributes).}




\subsubsection{Initial Coding Context Module}\label{Initial Coding based Context Module}


\qy{As mentioned in Section \ref{sec:Initial Coding}, we adopt RAHT for initial coding and represent the process of RAHT with a tree structure. Here, we exploit the context information hidden in the initial coding stage by extracting context features from both low- and high-frequency tree nodes, and propose our initial coding context module.}

\qy{\textbf{Context from High-frequency Nodes.} We first start to process the information of high-frequency nodes. As mentioned in Secs. \ref{sec:raht_tree} and  \ref{Serialization}, the transmitted symbol stream is composed of high-frequency coefficients, and each coefficient has a corresponding high-frequency node. In light of the strong relationships between the high-frequency coefficients and nodes, our initial coding context module follows \cite{huang2020octsqueeze} to extract latent embedding $\mathbf{h}_j$ of each high-frequency node. In particular, we feed context information, which is obtained in the initial coding stage, to a MultiLayer Perceptron (MLP) to obtain $\mathbf{h}_j$. For a given high-frequency node, the context information contains the depth level, the weight (\textit{i.e.}, the number of child nodes), the low-frequency coefficient, and the attributes. Note that, all information is available during decoding. 
}

\begin{figure*}[t]
	\centering
	\includegraphics[width=1.0\linewidth]{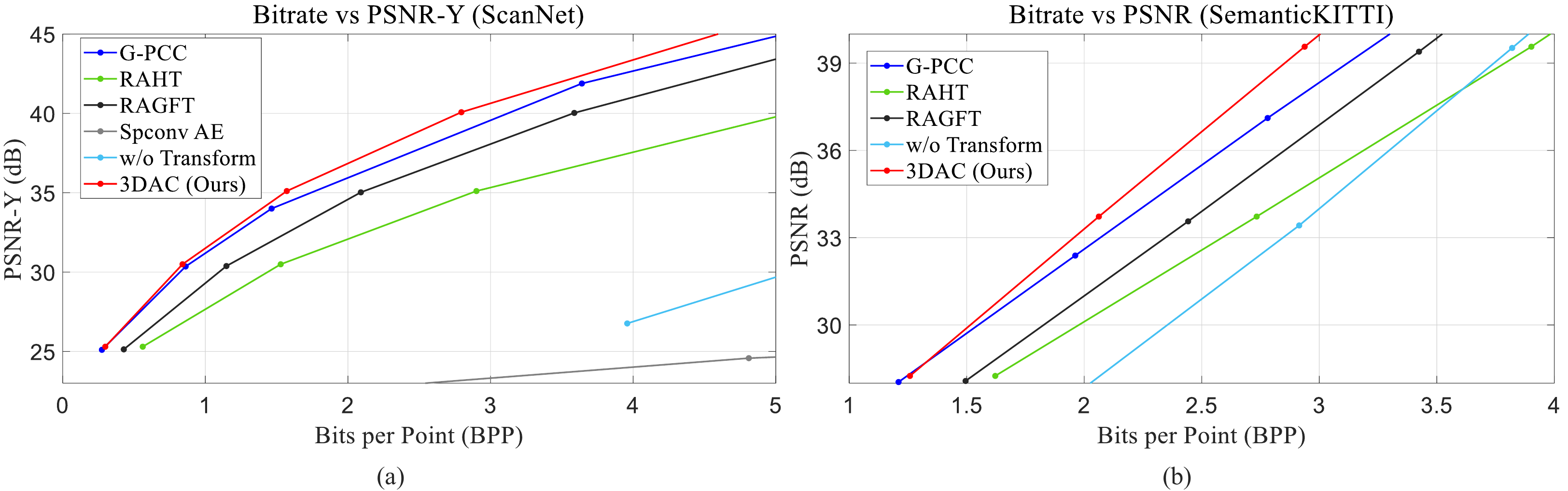}
	
    \caption{\qy{Quantitative results of different attribute compression approaches on the ScanNet (a) and SemanticKITTI (b) datasets.}}
	\label{fig:exp_scannet}
\end{figure*}


\qy{\textbf{Context from Low-frequency Nodes.} We further extract the information of low-frequency nodes, and fuse it with that of high-frequency ones into the initial coding context feature. Although low-frequency nodes do not interact directly with the transmitted symbols, there remains a massive quantity of information hidden in these nodes. Thus, we extract the context information that exists in these nodes. Here, we use SparseConv \cite{4dMinkpwski, tang2020searching} to process RAHT tree nodes for efficiency. In particular, \TBD{we perform 3D sparse convolutions at each depth level, and take both locations of low- and high-frequency nodes (\textit{i.e.}, center of corresponding subspace) as input points and their context information as input features. } With the progressive sparse convolution and downsampling, the context information of low-frequency nodes is fused with that of high-frequency nodes and diffused into multi-scale feature volumes. For each high-frequency node, we interpolate the latent features from multi-scale feature spaces at its 3D location, and then concatenate them into a final embedding feature $\mathbf{l}_j$.
}

We concatenate the latent embeddings $\mathbf{h}_j$ and $\mathbf{l}_j$, and feed them into a MLP to obtain the initial coding context $\mathbf{I}_j$ for each transmitted coefficients:

\begin{equation}
\mathbf{I}_j=\mathrm{MLP}([\mathbf{h}_j, \mathbf{l}_j]).
\end{equation}


\subsubsection{Inter-Channel Correlation Module}\label{Inter Channel based Context Module}

In most cases, a point cloud contains multiple attribute channels (\textit{e.g.}, color, normal) and there is significant information redundancy between different attributes. Thus, we propose an inter-channel correlation module to exploit the inter-channel correlation.

\textbf{Inter-Channel Coefficient Correlation.} For an uncompressed coefficient $ r_{j}^{(i)}$, we first incorporate the previous encoded coefficients $\left\{r_{j}^{(1)}, \ldots, r_{j}^{(i-1)}\right\}$ to explore the inter-channel coefficient correlation. Specifically, we use the previous encoded coefficients as prior knowledge and feed them into a MLP to extract the latent feature $\mathbf{c}_{j}^{(i)}$:

\begin{equation}
\mathbf{c}_{j}^{(i)}=\mathrm{MLP}([r_{j}^{(1)}, \ldots, r_{j}^{(i-1)}]).
\end{equation}
Here, we simply concatenate all previously encoded coefficients and use them as the input of an MLP layer.

\textbf{Inter-Channel Spatial Correlation.} We further aggregate the encoded coefficients through point cloud geometry to benefit the probability estimation. A key observation is that, the attributes of other points are helpful in predicting those of a given point. Therefore, we utilize the spatial relationships via the RAHT tree by incorporating all high-frequency nodes through 3D sparse convolution. Note that, only high-frequency nodes are considered in this part because low-frequency nodes do not contain transmitted coefficients. More specifically, we perform 3D sparse convolutions at each depth level, and take locations of high-frequency nodes as input points and their encoded coefficients as input features. For each high-frequency node, we obtain the coefficient spatial embedding $\mathbf{s}_j^{(i)}$ by interpolating the latent feature in multi-scale feature space.

Similar to the aggregation of initial coding context features, we first concatenate $\mathbf{c}_{j}^{(i)}$ and $\mathbf{s}_j^{(i)}$, and then feed them into a MLP layer to obtain the prior channel embedding $\mathbf{C}_j^{(i)}$. For the first attribute channel, we set the embedding $\mathbf{C}_j^{(1)}$ as all zeros features for model consistency. Thus, we can finally model $q\left(r_{j}^{(i)} \mid \mathbf{I}_j, \mathcal{R}^{(i-1)}, \ldots, \mathcal{R}^{(1)}\right)$ as $q(r_{j}^{(i)} \mid \mathbf{I}_j, \mathbf{C}_j^{(i)})$.

\subsubsection{Probability Estimation}\label{Probability Estimation}

We aggregate two context features $\mathbf{I}_j$ and $\mathbf{C}_j^{(i)}$ through a MLP layer to obtain the final latent embedding. This embedding is further used to generate learnable parameters for a fully factorized density model \cite{balle2018variational}, which is able to model the probability distribution $q\left(r_{j}^{(i)} \mid \mathbf{I}_j, \mathcal{R}^{(i-1)}, \ldots, \mathcal{R}^{(1)}\right)$. 


\subsection{Entropy Coding}

For entropy coding, we adopt an arithmetic coder to obtain the final attributes bitstream. In the previous steps, we have already converted attributes to transform coefficients in Sec. \ref{raht}, serialized these coefficients to a symbol stream in Sec. \ref{Serialization}, and obtained the probabilities of symbols in Sec. \ref{Probability Estimation}. At the final entropy encoding stage, the transform coefficients and the estimated probabilities are passed to the arithmetic coder to generate a final attribute bitstream for further compression. During decoding, the arithmetic coder is capable of restoring the coefficients with the same probability produced by our deep entropy model.




\subsection{Learning}

\qy{During training, we adopt the cross-entropy loss for the  deep entropy model:}
\begin{equation}
\ell=-\sum_{i}\sum_{j} \log q(r_{j}^{(i)} \mid \mathbf{I}_j, \mathbf{C}_j^{(i)}).
\end{equation}
\qy{To reduce the bitrate of the final attributes bitstream, we approximate the estimated distribution $q(\mathcal{R})$ with the actual distribution $p(\mathcal{R})$  by minimizing the cross-entropy loss.}


\section{Experiments}
\label{Sec:Experiments}


\begin{figure*}[t]
	\centering
	\includegraphics[width=1.0\linewidth]{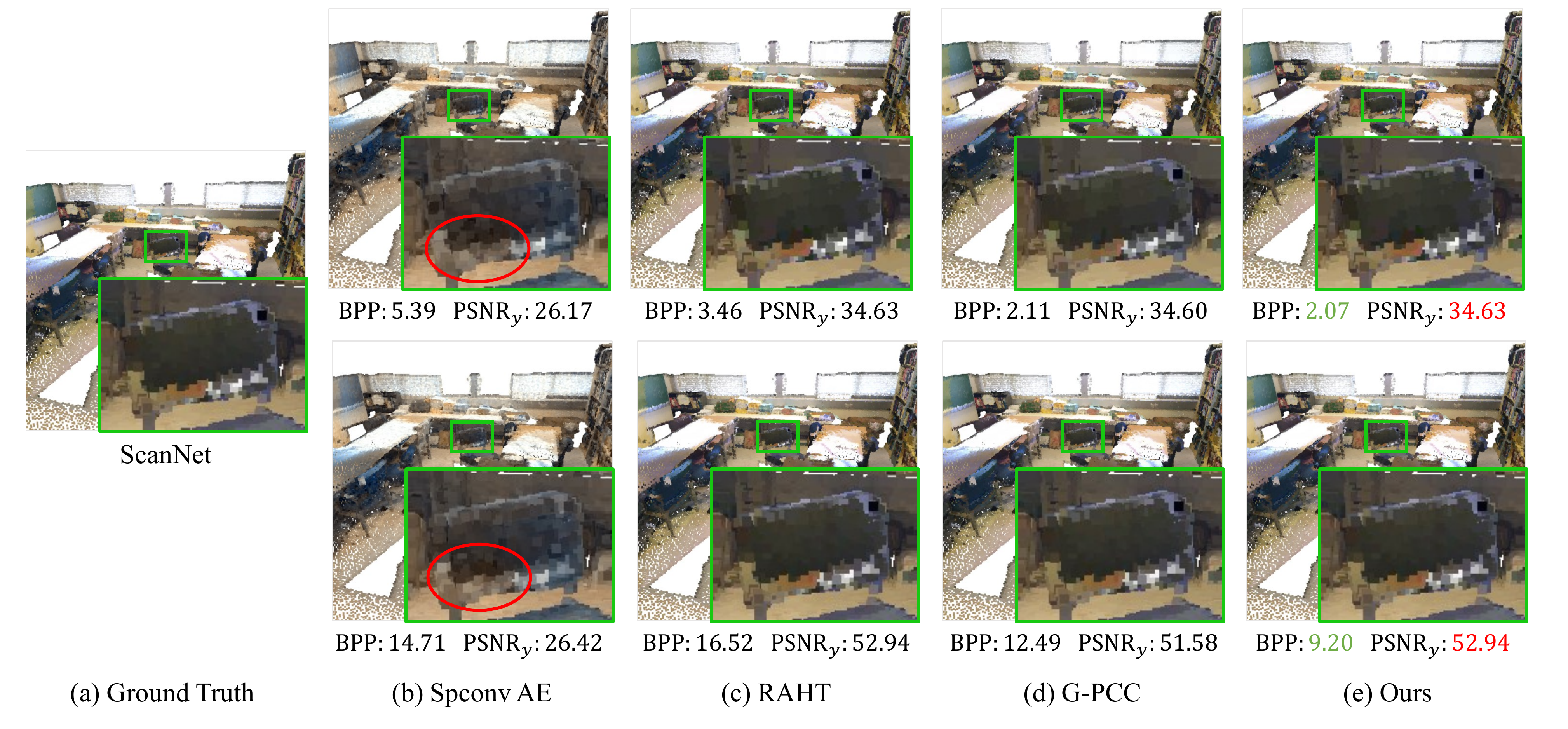}
	
    \caption{Qualitative results achieved by our \nickname{} and other baselines including Spconv AE, RAHT \cite{de2016compression}, and G-PCC \cite{schwarz2018emerging}. We visualize ScanNet scans at low and high bitrates, respectively. 
    Our method achieves the best compression quality (PSNR) with the lowest bitrates.}
	\label{fig:exp_qual_scannet}
\end{figure*}




In this section, we first evaluate the attribute compression performance of our \nickname{} on two point cloud datasets. Then, the effectiveness of our approach is validated on the downstream tasks. We finally conduct extensive ablative experiments to validate the contribution of each component.

\subsection{Experimental Setup}

\noindent\textbf{(1) Datasets.} We conduct experiments on the following two datasets:
\begin{itemize}[leftmargin=*]
\vspace{-0.2cm}
\setlength{\itemsep}{1pt}
\setlength{\parsep}{1pt}
\setlength{\parskip}{1pt}
  \item \textbf{ScanNet \cite{scannet}}. \qy{This is a large-scale indoor point cloud dataset containing 1513 dense point clouds. Each scan contains over ten thousand colored points. Following the official training/testing split, we use 1,201 point clouds for training and 312 point clouds for test. }
  \item \textbf{SemanticKITTI \cite{behley2019semantickitti}}. \qy{This is a large-scale outdoor LiDAR dataset with 22 sparse point cloud sequences. Point cloud reflectance captured by LiDAR sensors is used for attribute compression. Following the default setting in \cite{behley2019semantickitti}, we use 11 point sequences for training and the other 11 sequences for test.}
\end{itemize}

\noindent\textbf{(2) Evaluation Metrics.} \qy{The peak signal-to-noise ratio ($\rm PSNR$) is reported to evaluate the reconstruction quality. Following \cite{de2016compression}, we report the peak signal-to-noise ratio of the luminance component in ScanNet. Analogously,  the peak signal-to-noise ratio of reflectance is reported on the SemanticKITTI dataset. Bits Per Point ($\rm BPP$) is also adopted to evaluate the compression ratio.}

\noindent\textbf{(3) Baselines.} \qy{We compare the proposed \nickname{} with the following selected baselines:}
\begin{itemize}[leftmargin=*]
\vspace{-0.3cm}
\setlength{\itemsep}{1pt}
\setlength{\parsep}{1pt}
\setlength{\parskip}{1pt}
    \item \textbf{RAHT \cite{de2016compression}}. \qy{We use RAHT for initial coding and the run-length Golomb-Rice coder \cite{malvar2006adaptive} for entropy coding. }
    \item \textbf{RAGFT \cite{pavez2020region}}. \qy{This is an improved version of RAHT with graph transform. The run-length Golomb-Rice coder is also adopted for entropy coding. }
    \item \textbf{G-PCC \cite{schwarz2018emerging}}. \qy{This is a standard point cloud compression method (\textbf{G-PCC})  provided by MPEG\footnote{We use version 13.0 of G-PCC reference implementation: https://github.com/MPEGGroup/mpeg-pcc-tmc13}.}
    \item \textbf{Spconv AE}. \qy{We adopt torchsparse \cite{tang2020searching} to construct MinkUNet \cite{4dMinkpwski} for attribute reconstruction and use a fully factorized density model \cite{balle2018variational} for entropy coding. }
    \item \textbf{w/o Transform}. \qy{This is motivated by MuSCLE \cite{biswas2020muscle}. The attributes are transmitted without any transform and entropy coded by a probability density model \cite{balle2018variational} with geometric information. In contrast to \cite{biswas2020muscle}, we extract geometric context features through 3D sparse convolution from the current point cloud frame.}
\end{itemize}
Note that, uniform quantization of transform coefficients is used in our method, RAHT and RAGFT, while adaptive quantization is used in G-PCC.

\textbf{Implementation Details.} 
In order to simulate the real condition of point cloud compression, we voxelize the raw point cloud data with the 9-level and 12-level octree for ScanNet and SemanticKITTI, respectively, and assume that the geometry of point clouds has been transmitted separately. For ScanNet, a conversion is performed from the RGB color space to the YUV color space following the default setting of G-PCC \cite{schwarz2018emerging}. We adopt both of the initial coding context module and the inter-channel correlation module. For SemanticKITTI, considering that a single channel of attribute (\textit{i.e.}, reflectance) is included, we only use the initial coding context module and simply disable the inter-channel correlation module.





\subsection{Evaluation on Public Datasets}

\textbf{Evaluation on ScanNet.} The quantitative attribute compression results of different approaches on ScanNet are shown in Fig. \ref{fig:exp_scannet}. It can be seen that transform-based methods (\textit{i.e.}, 3DAC, G-PCC, RAHT, and RAGFT) significantly outperform other methods (\textit{i.e.}, w/o Transform and Spconv AE), demonstrating the effectiveness of the initial coding scheme. Additionally, the proposed \nickname{} consistently achieves the best results compared with other baselines. Although the same initial coding is adopted for our \nickname{} and RAHT, our method can achieve the same reconstruction quality with a much lower bitrate. This can be mainly attributed to the proposed entropy model and the learning framework. We also provide the qualitative comparison on ScanNet in Fig. \ref{fig:exp_qual_scannet}. It can be seen that our approach achieves better reconstruction performance even compared with the standard point cloud compression algorithm, G-PCC.

\textbf{Evaluation on SemanticKITTI.} \qy{The quantitative result of the compression performance achieved by different methods on the SemanticKITTI dataset is shown in Fig. \ref{fig:exp_scannet}. It is clear that the proposed \nickname{} consistently outperforms other methods by a large margin. This is primarily because that the proposed module is able to implicitly learn the geometry information from sparse LiDAR point cloud through 3D convolutions. We also noticed that existing compression baselines (including RAHT, RAGFT, and G-PCC) achieve unsatisfactory performance on this dataset, since these methods are mainly developed for dense point clouds. In contrast, the proposed \nickname{} is demonstrated to work well on both indoor dense point clouds and outdoor sparse LiDAR point clouds.}

\begin{figure}[t]
	\centering
	\includegraphics[width=1.0\linewidth]{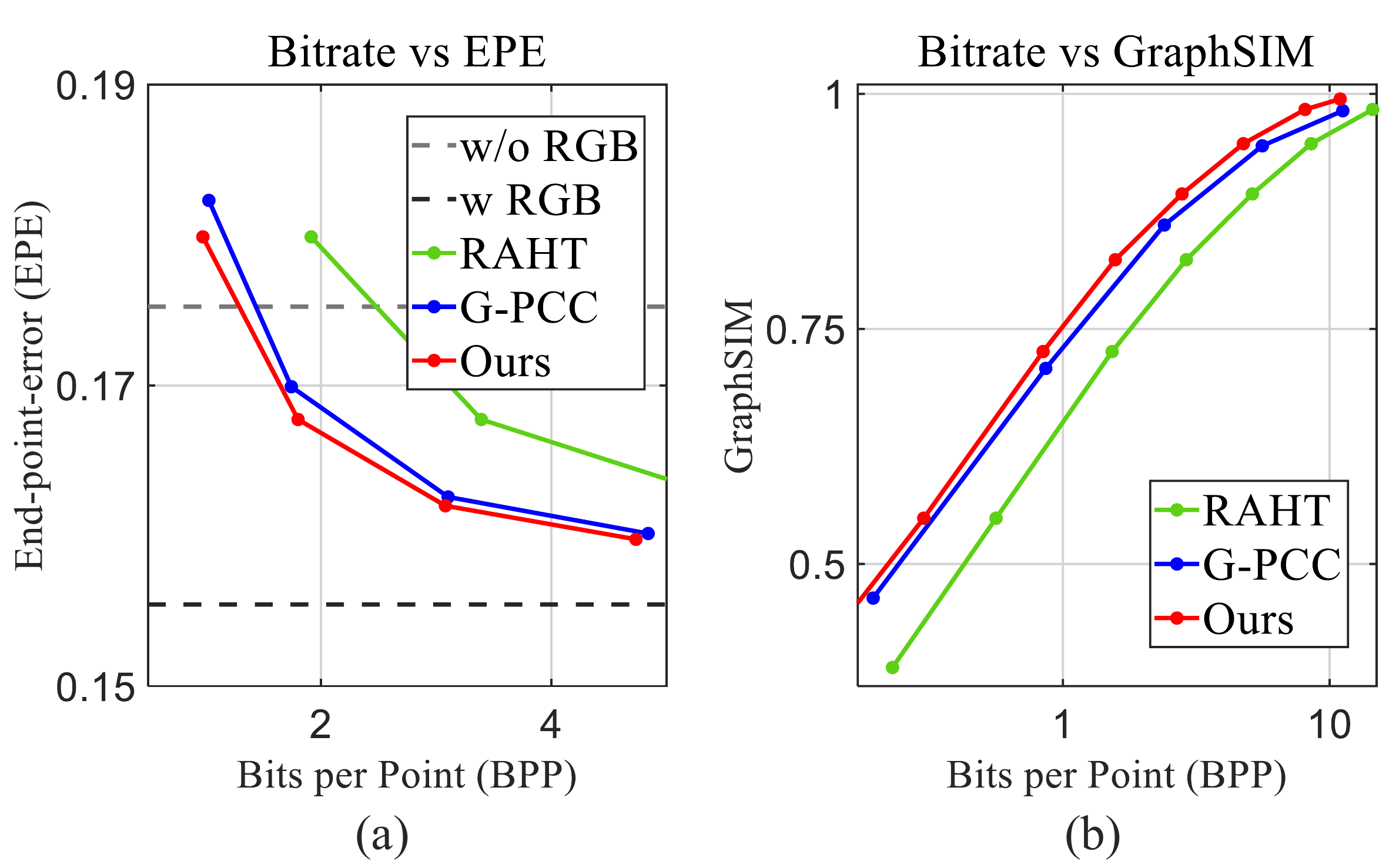}
	
    \caption{ The quantitative results on two downstream tasks. (a): Scene flow estimation on FlyingThings3D. (b): Quality asessment on ScanNet.
    }
	\label{fig:scannet_seg}
\end{figure}

	

\subsection{Evaluation on Downstream Tasks}

\textcolor{black}{We further evaluate the performance of attribute compression through two representative downstream tasks, including a 3D scene flow estimation task for machine perception and a quality assessment task for human perception.}

\textcolor{black}{\textbf{Scene Flow Estimation.} We present the performance of scene flow estimation on FlyingThings3D with FlowNet3D \cite{liu2019flownet3d}. In particular, we first train the network on the raw point clouds. Then, we evaluate the flow estimation performance on the compressed point clouds. The results of raw point clouds with and without attributes are also reported as `w RGB' and `w/o RGB', respectively. End-point-error (EPE) at different bits per point is adopted as the evaluation metric. As shown in Fig. \ref{fig:scannet_seg}, the proposed \nickname{} achieves better performance than RAHT and G-PCC. The performance on this downstream task further demonstrates the superior compression quality of our \nickname{}.}



\qy{\textbf{Quality Assessment.} For human perception tasks, we adopt point cloud quality assessment here since the human visual system is more sensitive to attributes. In particular, we adopt the GraphSIM \cite{yang2020inferring} metric to indicate the quality of the compressed point clouds, and higher GraphSIM score means lower attribute distortion. The scores at different bits per point are reported. As shown in Fig. \ref{fig:scannet_seg}, the proposed method achieves the highest GraphSIM score at the same bitrate, further demonstrating the effectiveness of our approach on downstream human vision-oriented tasks.
}


\begin{table}[htb]
\small 
  \centering
  \resizebox{0.35\textwidth}{!}{
    \begin{tabular}{cccc|c}
    \Xhline{2.0\arrayrulewidth}
  
    \multicolumn{2}{c|}{Initial Coding} & \multicolumn{2}{c|}{Inter Channel} & Birate  \\
   
    H & \multicolumn{1}{c|}{L} & C & S &  \\
    \hline
             &       &       &     & 3.85  \\
$\checkmark$ &       &       &           & 3.48  \\
$\checkmark$ & $\checkmark$ &          &       & 3.28  \\
$\checkmark$ & $\checkmark$ & $\checkmark$ &             & 3.08  \\
$\checkmark$ & $\checkmark$ & $\checkmark$ & $\checkmark$      & 2.79  \\
          \Xhline{2.0\arrayrulewidth}
    \end{tabular}%
    }
\caption{\qy{Ablation study on the initial coding context and inter-channel correlation modules. H, L, C, and S denote the information from High-frequency nodes, information from Low-frequency nodes, inter-channel Coefficient dependence, and inter-channel Spatial dependence, respectively.}}
  \label{tab:ablation_module}%
\end{table}%


\begin{table}[htbp]
\small 
  \centering
  \resizebox{0.26\textwidth}{!}{
    \begin{tabular}{cccc|c}
    \Xhline{2.0\arrayrulewidth}
    D & W & F & A & Birate \\
    \hline
             &       &       &     & 3.85  \\
$\checkmark$ &       &       &           & 3.64  \\
$\checkmark$ & $\checkmark$ &          &       & 3.57  \\
$\checkmark$ & $\checkmark$ & $\checkmark$ &             & 3.51  \\
$\checkmark$ & $\checkmark$ & $\checkmark$ & $\checkmark$      & 3.48  \\
          \Xhline{2.0\arrayrulewidth}
    \end{tabular}%
    }
    
\caption{\qy{Ablation study on information from high-frequency nodes. D, W, F, and A stand for the node’s RAHT tree depth level, weight, low-frequency coefficients, and attributes, respectively.}}   
  \label{tab:ablation_input}%
\end{table}%

\subsection{Ablation Study}
\qy{To further determine the contribution of each component in our framework, we conduct several groups of ablation experiments in this section. Note that, all experiments are conducted on ScanNet and all uniform quantization parameters are set as 10 for the same reconstruction quality.  First, we evaluate the initial coding context and inter-channel correlation modules. As shown in Table \ref{tab:ablation_module}, we start with using a non-parametric, fully factorized density model \cite{balle2018variational}, and then add information from high- and low-frequency nodes (\textbf{H} and \textbf{L}, respectively), inter-channel coefficient dependence (\textbf{C}), and finally adopt inter-channel spatial dependence (\textbf{S}). It is clear that the progressive incorporation of the proposed components can lead to a lower bitrate.
}

\qy{We then report the ablation studies over the context information of the RAHT-tree node in Table \ref{tab:ablation_input}. By progressively incorporating the node's depth level (\textbf{D}), weight (\textbf{W}), low-frequency coefficients (\textbf{F}), and attributes (\textbf{A}) into the context information of the high-frequency nodes, we can see a steady reduction of the encoding entropy, verifying the effectiveness of the proposed components.
}

\section{Conclusion}

In this paper, we presented a point cloud attribute compression algorithm. Our method includes an attribute-oriented deep entropy model considering both attribute initial coding and inter-channel correlations to reduce the storage of attributes. We showed the compression performance of our method over both indoor and outdoor datasets, and the results demonstrated that our approach has superior capability to reduce the bitrate as well as ensuring the reconstruction quality.



\smallskip\noindent\textbf{Acknowledgements.} This work was partially supported
by the National Natural Science Foundation of China (No. U20A20185, 61972435, 61971282), the Shenzhen Science and Technology Program (No. RCYX20200714114641140), and the Natural Science Foundation of Guangdong Province (2022B1515020103). Qingyong Hu was also supported by China Scholarship Council (CSC) scholarship and the Huawei AI UK Fellowship.


\clearpage

{\small
\bibliographystyle{ieee_fullname}
\bibliography{egbib}
}

\newpage
\appendix

\section*{Appendix} 

In this document, we describe details of Region Adaptive Hierarchical Transform (RAHT) and then describe network architecture details of our algorithm. We provide additional experiments on MPEG/JPEG point cloud compression database to validate the effectiveness of our approach, compare our method with learning-based baselines and also benchmark the runtime of our method to show its potential for real-world applications. Moreover, we show additional qualitative results on ScanNet and SemanticKITTI. Finally, we discuss the limitations and broader impact of our approach.

\section{Additional RAHT Details}

RAHT is a variation of Haar wavelet transform tailored for 3D point clouds. It first voxelizes point clouds and transforms point cloud attributes to low- and high-frequency coefficients along three dimensions repeatedly (\egours along the $x$ axis first, then the $y$ axis and the $z$ axis) until all points are merged to the entire 3D space. Here, we show the details about transforming two low-frequency coefficients to low and high-frequency coefficients.

Two neighboring points are merged during encoding and low- and high-frequency coefficients are generated from the corresponding low-frequency coefficients with the following transform:
\begin{equation}
\left[\begin{array}{l}
l_{d+1, x, y, z} \\
h_{d+1, x, y, z}
\end{array}\right]=\mathbf{T}_{w_{1}, w_{2}}\left[\begin{array}{c}
l_{d, 2 x, y, z} \\
l_{d, 2 x+1, y, z}
\end{array}\right]\label{con:raht_appendix},
\end{equation}
where $l_{d, 2 x, y, z}$ and $l_{d, 2 x+1, y, z}$ are low-frequency coefficients of two neighboring points along the $x$ dimension, and $l_{d+1, x, y, z}$ and $h_{d+1, x, y, z}$ are the decomposed low-frequency and  high-frequency coefficients. For the first depth level, point cloud attributes are regarded as low-frequency coefficients. Here, $\mathbf{T}_{w_{1}}$ is defined as
\begin{equation}
\mathbf{T}_{w_{1}, w_{2}}=\frac{1}{\sqrt{w_{1}+w_{2}}}\left[\begin{array}{cc}
\sqrt{w_{1}} & \sqrt{w_{2}} \\
-\sqrt{w_{2}} & \sqrt{w_{1}}
\end{array}\right],
\end{equation}
where $w_{1}$ and $w_{2}$ are the weights (i.e., the number of leaf nodes) of $l_{d, 2x, y, z}$ and $l_{d, 2x+1, y, z}$, respectively. Low-frequency coefficients are directly passed to the next level if the point does not have a neighbor.





\section{Additional Architecture Details}


\smallskip\noindent\textbf{RAHT tree node context.} For a given RAHT tree node, the context information contains the depth level, (\textit{i.e.}, depth of the node in the RAHT tree), the weight $w$, (\textit{i.e.}, the number of child nodes), the reconstructed low-frequency coefficients $l$ (\textit{i.e.}, the accessible low-frequency coefficients during decoding) and the reconstructed attributes $a$, (\textit{i.e.}, mean attributes of all points in the corresponding subspace). Note that the reconstructed attributes can be obtained by $a=\frac{l}{\sqrt{w}}$.

\smallskip\noindent\textbf{Architecture Details.} For context feature extraction from high-frequency nodes and inter-channel coefficient correlation, we use a 3-layer MLP (8, 16 and 8 dimensional hidden features) with context of high-frequency nodes and previous encoded coefficients as input, respectively. For context feature extraction from low-frequency nodes and inter-channel spatial correlation, we adopt torchsparse \cite{tang2020searching} to construct 4 3D sparse convolution layers (3, 3, 6 and 8 dimensional hidden features, and convolution stride as 2) and use trilinear interpretation to obtain latent features from output feature volume of each convolution layer. Besides, latent feature aggregation is realized by a 3-layer MLP (8, 16 and 8 dimensional hidden features) for both initial coding context module and inter-channel correlation module.




\smallskip\noindent\textbf{Implementation Details.}
Our network is implemented in PyTorch and trained with one NVIDIA 1080TI GPU. We train our model over 20 epochs using the Adam optimizer with an initial learning rate of 0.01.



  
  
  
  

\begin{figure*}[t]
	\centering
	\includegraphics[width=1.0\linewidth]{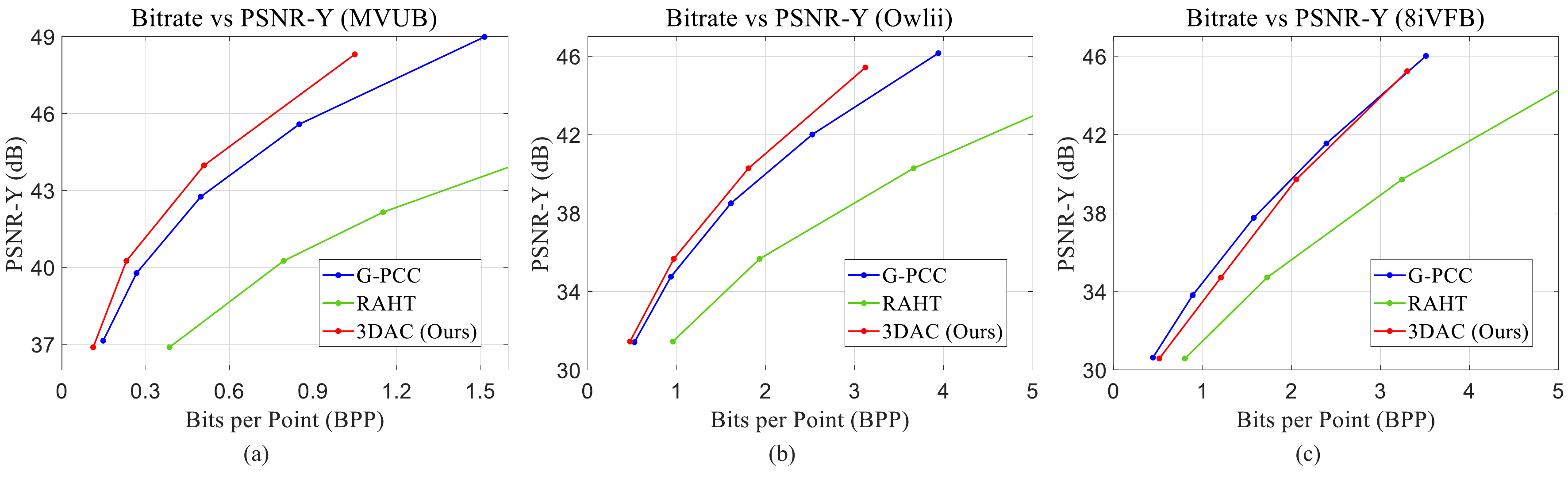}
	
    \caption{Additional experiments on MPEG/JPEG database. Quantitative results of different attribute compression approaches on the MVUB (a), Owlii (b) and 8iVFB (c) datasets.}
	\label{fig:mpeg_pcc_suppl}
\end{figure*}

\section{Additional Experiments}

\smallskip\noindent\textbf{MPEG/JPEG database.} We also conduct experiments on MPEG/JPEG point cloud compression datasets, including MVUB \cite{loop2016microsoft}, Owlii \cite{xu2017owlii} and 8iVFB \cite{d20178i}. All these datasets contain four to five dynamic human point cloud sequences. We voxelize point clouds of these dataset with a 9-level octree. For MVUB, we use subjects, Andrew, David, Phil and Ricardo for training and Sara for testing. For Owlii, we use basketball player, dancer, exercise for training and model for testing. For 8iVFB, we use longdress, loot, redandblack for training and soldier for testing.


The quantitative attribute compression results on MPEG/JPEG database are shown in Fig. \ref{fig:mpeg_pcc_suppl}. Our 3DAC achieves better compression performance with other baselines on MVUB and Owlii. Due to the huge diversity of attributes in 8iVFB, our method has a small performance gap compared with the standard point cloud compression software, G-PCC, while it is still much better than RAHT. The results on these point cloud compression datasets further illustrate the effectiveness of our method.

\begin{figure}[h]
    \vspace{-0.2cm}
	\centering
	\includegraphics[width=1\linewidth]{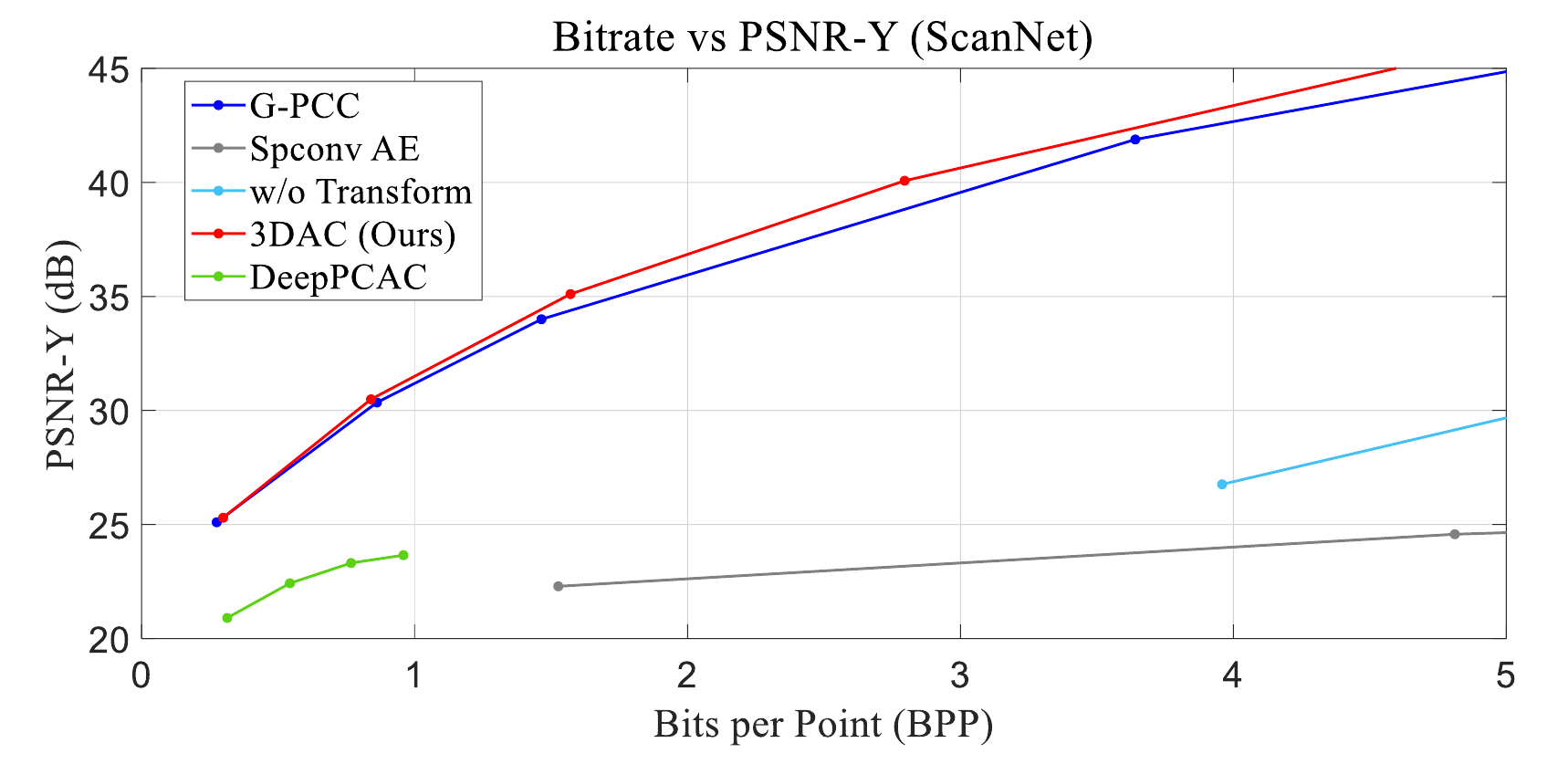}
	\vspace{-0.8cm}
    \caption{Quantitative results of different learning-based attribute compression approaches on the ScanNet dataset.}
	\label{fig:exp_quant_dl}
	\vspace{-0.3cm}
\end{figure}

\smallskip\noindent\textbf{Comparison with learning-based methods.} In order to demonstrate the superiority of our method, we additionally compare our 3DAC with other complex learning-based baselines. In particular, we additionally include a concurrent work, DeepPCAC \cite{sheng2021deep}, on ScanNet. As shown in Figure \ref{fig:exp_quant_dl}, our 3DAC significantly outperforms all other learning-based baselines.

\smallskip\noindent\textbf{Run time.}
We benchmark runtime of our method on ScanNet with an Intel Core i7-8700 CPU and a Nvidia GeForce GTX 1060 6GB GPU. In our experiments, the arithmetic coder is implemented in C++, and the initial coding method and our entropy model are implemented in python. We set the quantization parameter as 10. The encoding and decoding time of our method are \TBD{3.21s and 3.27s}, respectively, and those of G-PCC \cite{schwarz2018emerging}, which is implemented in C++, are 0.36s and 0.31s. Although our method is slower than G-PCC, we believe that it is possible to speed up our method with parallel computation and an optimization in data I/O.


\section{Additional Qualitative Results}

\begin{figure*}[t]
	\centering
	\includegraphics[width=1.0\linewidth]{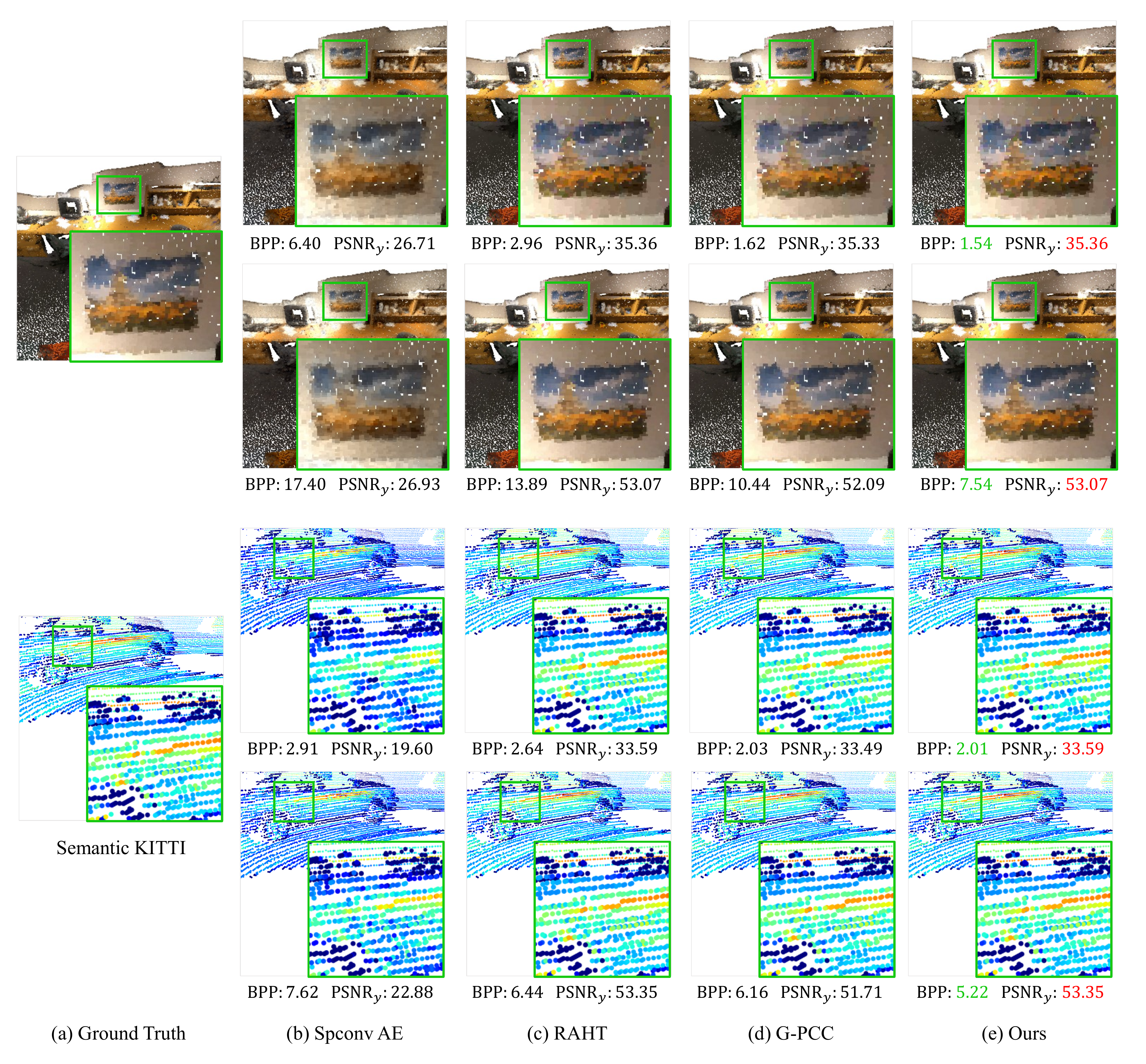}
	
    \caption{Additional qualitative results achieved by our method and other baselines including Spconv AE, RAHT \cite{de2016compression} and G-PCC \cite{schwarz2018emerging}. We visualize ScanNet scans with RGB colors and Semantic KITTI with the intensity of reflectance at relatively low and high bitrates. It is clear that our method can achieve the best compression quality (PSNR sores) with the lowest bitrates.
    }
	\label{fig:qual_suppl}
\end{figure*}

We show additional qualitative results on ScanNet and SemanticKITTI to show our compression performance in Fig. \ref{fig:qual_suppl}. As shown in the figure, our method can retain better reconstruction quality as well as reducing bitrates.


\section{Limitations and Broader Impact}

In the current experimental setup, our algorithm only achieves lossy point cloud compression. In specific, we voxelize point clouds with an octree (which leads to geometry distortion), and then adopt RAHT and uniform quantization to process the voxlized point clouds (which leads to attribute distortion). Due to these operations, our framework can not realize lossless compression. A possible solution is to upsample compressed attributes form voxelized points to original points through interpolation, and then transmit the residual of attributes. We leave the attribute interpolation and the residual coding as a future study.

Our point cloud attribute compression algorithm directly helps to 3D data compression, storage and transmission. Thus, we do not foresee any direct negative societal impact of our method. However, the compression and transmission of point cloud data, such as human body and human face, may indirectly lead to invasion of privacy. Thus, we need to be aware of some malicious applications of our method.

\end{document}